# A dynamic Bayesian optimized active recommender system for curiosity-driven "Human-in-the-loop" automated experiments.

Arpan Biswas,[1*] Yongtao Liu[1], Nicole Creange[1†], Yu-Chen Liu[2], Stephen Jesse,[1] Jan-Chi Yang[2], Sergei V. Kalinin[3], Maxim A. Ziatdinov,[1,4] Rama K. Vasudevan[1‡]

[1] Center for Nanophase Materials Sciences, Oak Ridge National Laboratory, Oak Ridge, TN 37831

[2] Department of Physics, National Cheng Kung University, Tainan 70101, Taiwan

[3] Department of Materials Science and Engineering, University of Tennessee, Knoxville

[4] Computational Sciences and Engineering Division, Oak Ridge National Laboratory, Oak Ridge, TN 37831

**Abstract**

Optimization of experimental materials synthesis and characterization through active learning methods has been growing over the last decade, with examples ranging from measurements of diffraction on combinatorial alloys at synchrotrons, to searches through chemical space with automated synthesis robots for perovskites. In virtually all cases, the target property of interest for optimization is defined *apriori* with limited human feedback during operation. In contrast, here we present the development of a new type of human in the loop experimental workflow, via a Bayesian optimized active recommender system (BOARS), to shape targets on the fly, employing human feedback. We showcase examples of this framework applied to pre-acquired piezoresponse force spectroscopy of a ferroelectric thin film, and then implement this in real time on an atomic force microscope, where the optimization proceeds to find symmetric piezoresponse amplitude hysteresis loops. It is found that such features appear more affected by subsurface defects than the local domain structure. This work shows the utility of human-augmented machine learning approaches for curiosity-driven exploration of systems across experimental domains.

**Keywords**: Human-in-the-loop experiments, automated experiments, Bayesian optimization, active recommendation system, dynamic target selection, deep kernel learning.

---

[*] biswasa@ornl.gov
[†] Current affiliation: Tjoapack LLC, Clinton, TN 37716
[‡] vasudevanrk@ornl.gov

# Introduction

The achievable progress in the field of automated and autonomous experiments, and the idea of 'self-driving' laboratories more generally, hinges on the ability of probabilistic machine learning models to be used to rapidly identify areas of the parameter space that have a high (modeled) likelihood of optimizing target properties of interest.[1–5] Recent examples include explorations of chemical space[6] in the synthesis of nanoparticles[7] and thin films for photovoltaic applications,[8,9]. Additionally, numerous examples exist of autonomous microscopes that can be used to identify structure-property relationships in both electron[4] and scanning probe spectroscopies,[10,11] as well as scattering measurements at the beamline, for e.g., efficient capture of diffraction patterns for phase mapping or for strain imaging.[12–15] Such work seeks to improve not only the efficiency at which the target property of interest can be found and/or maximized, but also to improve our understanding of how composition and structure impact functionality, ideally unearthing them in real-time.

In nearly all cases of active learning within experiments, the target property of interest is defined *apriori*. This target can be human-designed behavior of interest, for example some measured property, or feature of a spectra that is captured such as area, peak position, peak ratio, etc. In these cases, the objective of the experiment is to efficiently probe the parameter space to maximize the selected target. Alternatively, an information-theory approach can be used where the goal is instead to minimize uncertainty of a developed surrogate model. In both cases however, the human is generally kept out of the loop after the target is selected and a sampling policy is initiated. Indeed, a celebrated review of Bayesian optimization is titled "*Taking Humans out of the Loop: A Review of Bayesian Optimization*".[16]

However, this may not always be ideal. In some situations, experimentalists would prefer to observe a few of the spectra prior to target formulation, to obtain a sense of the potential novelty of the regions that could be probed. Additionally, with little prior knowledge, it may also be challenging to design a suitable scalarizer that captures the essence of the target. Moreover, it is possible that the human may prefer one target on seeing some spectra, and then observe something more interesting in subsequent points and decide that is more worthwhile exploring. This dynamic setting and changing of targets, that need to be inferred by the algorithm, is a problem is well encountered in other fields such as social media, and has been solved via recommendation engines, which are built from user voting ('likes') to populate the feeds with content agreeable with the user.[17–19]

**Human-in-the loop based Automated Experiment (AE) Workflow**

Here we present the development of a new method of automated experiment which employs a human in the loop experimental workflow, which we term the Bayesian optimized active recommender system (BOARS). We develop and apply it to the case of finding spectra encountered during piezoresponse force spectroscopy measurements, first trialing the method on pre-acquired data to gauge the effectiveness, and then implementing it in real time on an operating instrument. The framework allows the human operator to vote for a certain number of spectra to construct a target, and then proceeds to explore the search space optimally in view of retrieving spectra that bear a strong structural similarity to the target, and in the process, unearth key structure-property relationships present autonomously. In this manner, we bypass the need for a pre-defined target and add flexibility to a standard automated experiment, where rather than fixing a target prior to the start of the experiment, the human operator retains the ability to dynamically adjust the target via real-time result assessments.

The overall framework has two major architectural components - an active recommender system (ARS) and Bayesian Optimization (BO) engine. The ARS is developed as a dynamic, human-augmented computational framework where, given a location in the search area of the material samples, the microscope performs a spectroscopic measurement at the location in real time. This spectrum provides knowledge about various key features (e.g., it can be energy loss, nucleation barriers, degree of crystallinity, etc.). The ARS system allows the user to upvote and downvote spectra according to the features of their own interest, and this method is free from any generalized objective functions. Previously, human-augmented recommender systems have been developed in microscopy in accelerating meaningful discoveries in different field of applications such as rapid validation of thousands of biological objects or specimen tracking results,[20] and rapid material discovery of novel lithium ion conducting oxides through synthesis of unknown chemically relevant compositions.[21]

The second part of the architecture is the BO engine, which guides the path to locate the regions of interest with maximum structural similarity to the human-upvoted spectra, through sequential updating with a computationally cheap surrogate model and enable an efficient trade-off between exploration and exploitation of the unknown search area. Bayesian optimization (BO) or (multi-objective) BO,[16,22–26] has been originally developed as a low computationally cost global optimization tool for design problems having expensive black-box objective functions. BO has been extensively applied for rapid exploration of large material[27–34] and chemical[35,36] control parameters and/or functional properties space exploration to enable optimization towards desired device applications. Here, the BO replicates the expensive function evaluations with a cheap (scalable) surrogate model and then utilizes an adaptive sampling technique through maximizing an acquisition function to learn or update the knowledge of the parameter space towards finding the optimal region. Over the years, development of BO has been extended for various complex problems. Biswas and Hoyle extended the application of BO over discontinuous design space by remodeling into a domain knowledge driven continuous space[37]. BO has been extended in discrete space such as in consumer modeling problems where the responses are in terms of user preference discrete[38–41]. Here, Thurstone[39] and Mosteller[40] transform the user preference discrete response function into continuous latent functions using Binomial-Probit model for binary choices, whereas Holmes[41] uses a polychotomous regression model to applicable for more than two discrete choices. For practical implementation of BO over high-dimensional input space, some examples like Dhamala et.al.[42], Valetti et. al.[43] and Wang et. al.[44] attempted the approach of random embedding in a low-dimensional space; Grosnit et. al.[45] and Biswas et. al.[46] attempted the approach to project into a low-dimensional latent space with variational autoencoder; and Oh et. al.[47], Wilson et. al.[48] and Ziatdinov et.al.[49] tackles with implementing special kernel functions.

A Gaussian Process Model (GPM)[50] is generally integrated in BO as the surrogate model, which also provides the measure of uncertainty of the estimated expensive functions over the parameter space such that the uncertainty is minimal at explored regions and increases towards the unexplored regions. Alternatively, random forest regression has also been proposed as an expressive and flexible surrogate model in the context of sequential model-based algorithm configuration.[51] The detailed workflow of BO and mathematical representation of GPM is provided in Supplementary Material (Appendix A). Once a cheap surrogate model is fitted in a BO iteration with the sampled data, the next task is to find the next best locations for sampling through maximizing the acquisition function (AF). The latter defines the likelihood of finding the region of interest or better objective function values. Several acquisition functions, such as Probability of Improvement (PI), Expected Improvement (EI), Confidence Bound criteria (CB) have been developed with different trade-offs between exploration and exploitations.[52,53,22,54]

In all the stated BO applications where the target is required to be set prior to the optimization, in this paper, the proposed approach bypasses that requirement by introducing a human-in-the loop architecture, thus adding flexibility to the automated experimental workflow. We additionally explore the effect of local structure encoded in image patches and different kernel functions on the performance of the optimization trajectory.

Our main research contributions are:

a) We introduce a human-in-the-loop approach within a BO-based AE workflow to add flexibility to the experimentalist to define a target spectrum, representing several material properties for a given application, after going through visual inspections and assessments of captured spectra. We term this a "*Active Recommender System (ARS)*".
b) We trial this architecture on pre-acquired data to simulate the process on a real sample, to determine the effects of different kernels on the overall optimization trajectory
c) We use this knowledge to implement the full workflow, in real-time, on an operating microscope, showing the example of attempting to find symmetric switching properties in ferroelectric thin films via upvoting symmetric-looking hysteresis loops and downvoting others.

**Bayesian Optimized Active Recommender Engine (BOARS)**

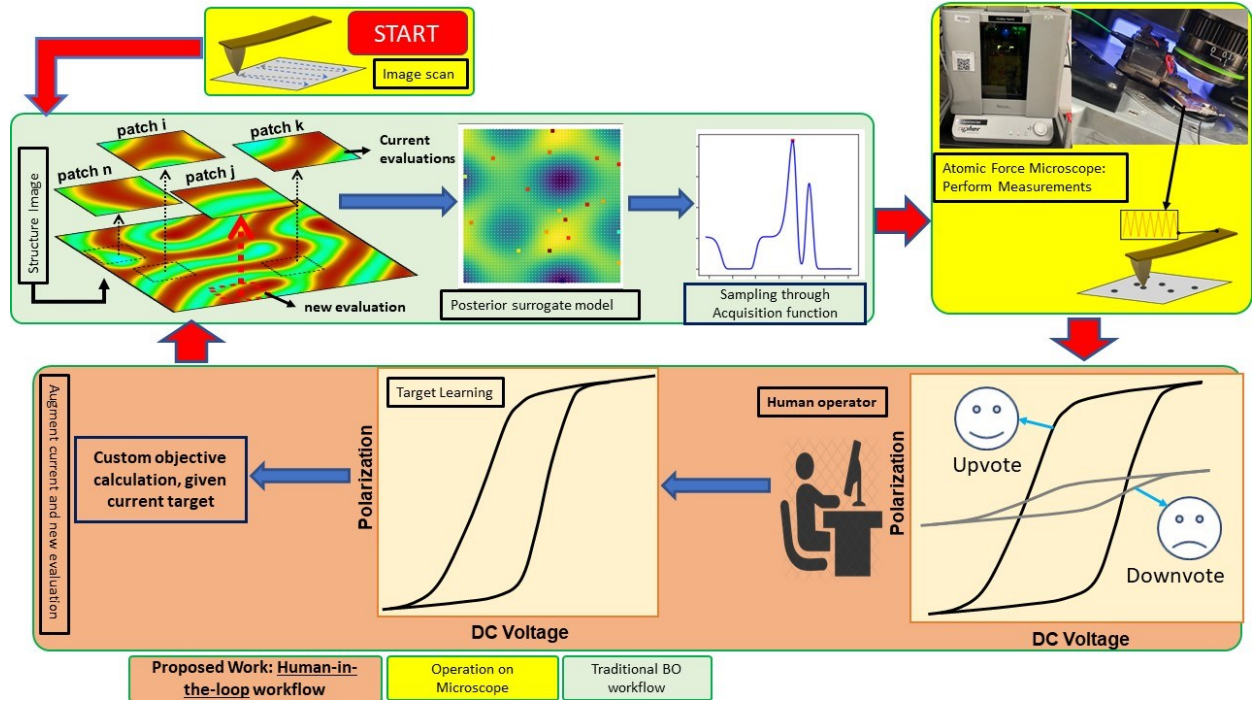

**Figure 1.** Dynamic, human-in-the-loop BOARS architecture. In this AE workflow, the step under orange region is the contribution in this paper where we introduce a human-operator active recommender system to vote and build a target spectral through visual inspection and define a reward-based structural similarity-based objective function. The steps under green and yellow regions are traditional Bayesian optimization (BO) workflow and instrument (microscope) operations to scan an image of the sample and capture spectra at BO guided locations over the image space. Additionally, the red highlighted arrow between yellow and orange region is another contribution of the paper which builds the connection of the workflow between human-operated tasks (recommender system) and the microscope operations for real-time implementations of this overall human-in-the-loop AE architecture. The other red highlighted arrows signify the coupling between different environments of the framework: between microscope and traditional BO workflow and vice-versa, and between human-in the loop part and traditional BO workflow.

    **Figure 1** shows the overall high-level structure of the BOARS system with the detail flow-chart of the algorithm is provided in Supplementary Materials (Appendix Fig B1). The workflow can be stated as follows: Given a material sample, we run the microscope to scan a high-resolution image, which is the parameter space for the exploration. Next, we segment the image space into several image patches of a set window size, $w$. We define this local image patches as the input for Bayesian optimization. Next, we initialize BO and capture spectra from microscope measurements at few randomly generated locations. Next, we introduce the steps for human-operation which is the major contribution of this work. Given a spectrum, the user (human) visualizes the spectra and provide subjective vote on its quality. As the user votes on spectra, a target is defined as $T_i = S_i$ where $S_i$ is the $i^{th}$ spectrum. With the next subsequent upvoting of other spectra, the target is

accordingly updated as per eq. 1 below. Once all the voting is complete for the first few randomly selected spectra, a human-guided objective function is calculated as eq. 2.

$$T_i = \left.\left((1-p_i) * \sum_{ii=1}^{i-1} v_{ii} * T_{i-1}\right) + (p_i * v_i * S_i)\right/ \left((1-p_i) * \sum_{ii=1}^{i-1} v_{ii}\right) + (p_i * v_i) \quad (1)$$

$$Y_i = \psi(T_j, S_i) + v_i * R \quad (2)$$

where $T_i$ is the target after $i^{th}$ spectra assessment given the user upvoted the spectra, $p_i$ is the user preference (0-1 with 1 being highest) of adding features of new spectra to the current target, $v_i$ is the user vote of the $i^{th}$ spectra, $Y_i$ is the objective function value for the $i^{th}$ spectra, $T_j$ is the target after voting all the $j$ spectra, $R$ is the reward on voting. The objective function is the voting augmented structural similarity index function where $\psi$ is the structural similarity function, computed from the function "*structural_similarity*" in "*skimage.metrics*" Python library.[55] Then, given the dataset with input local image patches and output objective function value, we run the BO—fitted with a Gaussian process model, and maximizing the acquisition function derived from the GP estimations. The acquisition function suggests the next best location to capture spectra. Next, microscopic measurement is carried out to retrieve the spectrum at the stated location and similar human assessment is carried out on the new spectrum. Given whether the user upvoted or downvoted the new spectrum, the target is either updated following eq (1) or remains the same, and the objective function is calculated following eq (2). This iterative GP training with new data, undertaking microscope measurements at new locations, and the introduced human-in-the-loop process to evaluate the spectra, update the target and calculate the objective function continues until the user is satisfied with the current target, which can be provided in a "Yes/No" popup message after every iteration. Then, the remaining iterations are carried out until BO convergence without any further human interaction, with the objective function value modified to eq. (3).

$$Y_{j+k} = \psi(T, S_k) \quad (3)$$

where $Y_{j+k}$ is the objective function for $k^{th}$ iteration of BO, after randomly sampling $j$ spectra. As seen, we removed the human-voting part as now the target $T$ is fixed and the task is to identify the spectra maximizing the structural similarity with the target. Thus, in the proposed design, within the loop of BO, here we define and refine the target (spectral structure) through human assessment, and simultaneously optimize either the human-augmented objective function or the fully automated objective function, following equation 2 or 3 respectively, given the state of decision making in updating the target. The detail mathematical algorithm of the methodology is provided later in the Method section for additional information.

## Results and Discussion

We first begin by testing the BOARS system on pre-acquired data (i.e., data where the ground truth is known, and not on the active microscope) to determine the applicability of the method and to note the effects of hyperparameters. To this aim we explored data from two PbTiO₃ (PTO) thin film samples. The samples are both 200nm-thick $PbTiO_3$ thin films grown on (110) $SrTiO_3$ via pulsed laser deposition, with 'designer' grain boundaries fabricated by a process outlined in ref[56]. In this instance, our measurements are not in the vicinity of the grain boundary; however, the domain structure of the PTO sample is dependent on the strain imparted by the thickness of the underlying substrate, and this leads to different domain structures for the part of the sample rotated with respect to the underlying (110) STO, as the underlying substrate is a rotated (110) STO membrane of limited (~10nm) thickness. As such, both samples imaged display different domain patterns enabling us to test the BOARS on samples with different domain features. For this paper, we refer to PTO sample 1 as the sample where the domains imaged are on the original (110) oriented STO crystal (thickness 500μm), and PTO sample 2 as the sample where we image the region of the sample where the sample is rotated (~2 degrees) with respect to the substrate and has a much lower thickness of the STO (and thus likely to be much less strained).

**Case Study: BOARS analysis with different kernels on existing PTO data**

To demonstrate the method, and before implementing it on the real-time microscope, we began with a full ground truth dataset where we measured the spectral data for all the grid locations (2500 grid points on a 50x50 grid).

To study the performance, we first considered the BOARS architecture with the Gaussian process model used as the surrogate model, and a standard periodic kernel function. It is to be noted we tested with other inbuilt kernel functions like radial basis, matern kernel, but periodic kernel provided superior exploration. The hyper-parameter of the kernel function is optimized with Adam optimizer[57] with learning rate = 0.1. We started with 10 initial samples, $j = 10$ and 200 BO iterations, $M = 200$, a total of 210 evaluations. In regard to incorporating the local image patches as additional channel for structure-spectra learning, we considered the image patch of window size, $w = 4\,px$. Thus, the dimension of each input, $X_1$, is an array of 16 elements. For a comparative study, we upvoted spectra that appeared (by eye) to possess roughly symmetrical hysteresis loops in terms of amplitude, i.e., similar remanent piezoresponse for positive and negative bias. For both PTO samples, we utilized voting (target learning) of the first 10 spectra and then fixed the target for the remaining iterations. The detailed user voting of the spectra used to set the final target for both PTO samples is provided in the Supplementary Material (see Figs. B2, B3). Figures 2 and 3 show the detailed analysis of the initial piezoresponse force microscopy image, as well as the structural similarity maps, after adaptive learning with BOARS system, for first and second PTO samples respectively.

Firstly, it can be clearly seen comparing the scanned images of the PTO samples (Figure 2(a), Figure 3(a)) with the respective structural similarity (ground truth) images (Figure 2(d), Figure 3(d)) that these are not highly correlated, particularly for Figure 3. That is, there is minimal correlation between the initial PFM scan and the structural similarity map. This is expected in cases where the features targeted in the spectral domain, here, symmetric remnant response, is not significantly dependent on the surface domain structure image and is likely to be more heavily determined more by sub-surface defects that are not manifest in the image. The objective for the appropriate model would be to balance between prior (local domain correlation) knowledge from scanned images and the posterior objective function knowledge through sequential learning, such that it tends towards efficient estimation of the structural similarity map at the explored and unexplored regions.

Here, inspecting Figure 2, as the BO converges with exhaustion of the sampling, we can see the majority of the user-desired (symmetrical) spectral locations are found (see figures (2. c (i), (iii)) with relatively few areas of non-satisfactory spectral response ((see figures (c (ii), (iv)), ultimately learning the boundaries between the desired and undesired spectral regions with explorations.

The interesting observation is that the GP estimated map (Fig. 2 (e)) is over mimicking the scanned image (Fig. 2 (c)), which shows the indication of possible overfitting to the input data or in other words, bias towards the prior knowledge rather than the balanced trade-off between inputs and output-defined objective functions. For example, if we focus on the region within the white circle in Fig. 2(c), the PFM amplitude image has low values (dark region) while the respective ground truth has high objective function value (red region). While the significant exploration is carried out in that region as expected, the uncertainty map (Fig. 2(f)) is unexpectedly estimated to have the most variability within this region, and relatively lower variance exists in unexplored locations. This could be due to the inefficient learning of traditional kernel functions over high dimensional inputs.[58] We also note even worse performance for Fig. 3,. Thus, we can see the traditional kernel does not have the capability to adjust the learning with gathering more information from the spectra.

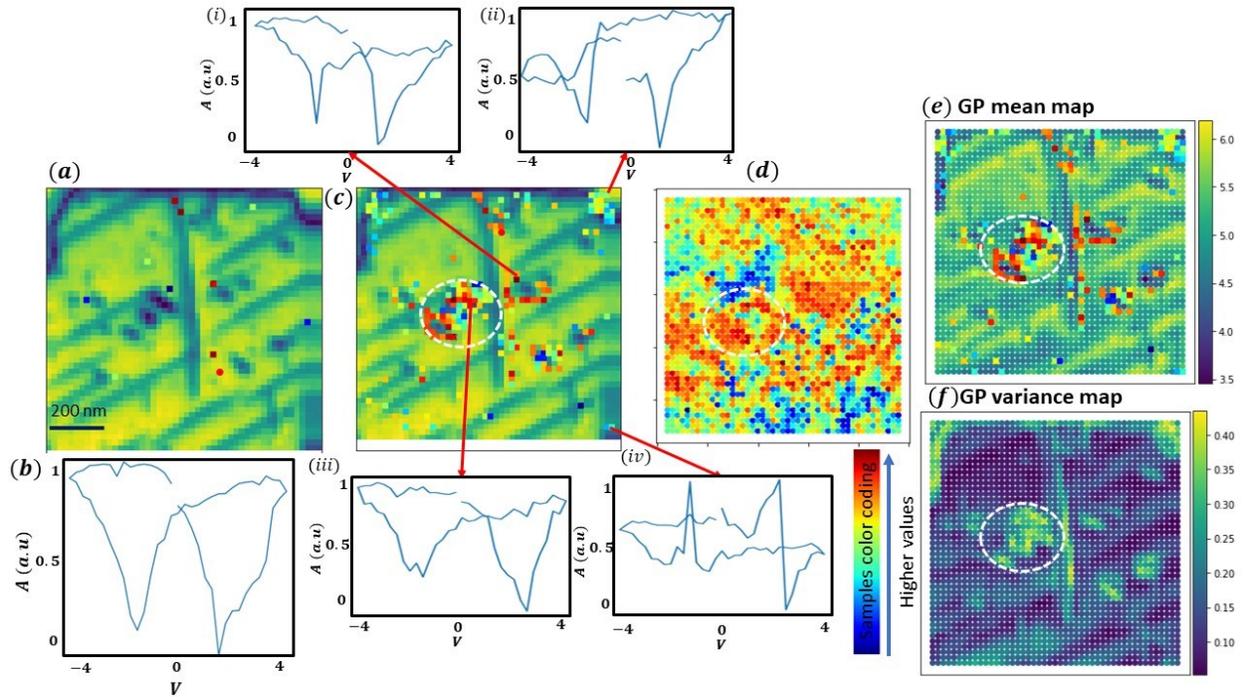

**Figure 2.** Analysis for PTO sample 1: BOARS with image patches. (a) Downsampled PFM amplitude image of the PTO film, with exploration points for the spectral locations where the user voted (b) final learned target spectral structure after voting through explored spectra in (a), (c) Plot of (a) with all the explored spectral locations overlaid (d) ground truth image, i.e., the structural similarity map as in eq. 3, given the user voted target spectra. (e) estimated structural similarity map from the surrogate model, with all the explored spectral locations, (f) map of the model's associated uncertainty. The color of the explored locations represents the human-augmented

objective function values in (a) and automated objective function values in (c), (e) as per the colormap provided. This color coding matches with the full ground truth image in (d). Within sub-figure (c), (i)-(iv) are the visualization of the spectra at some of the BO explored locations. Scale bar in (a) is 200 nm.

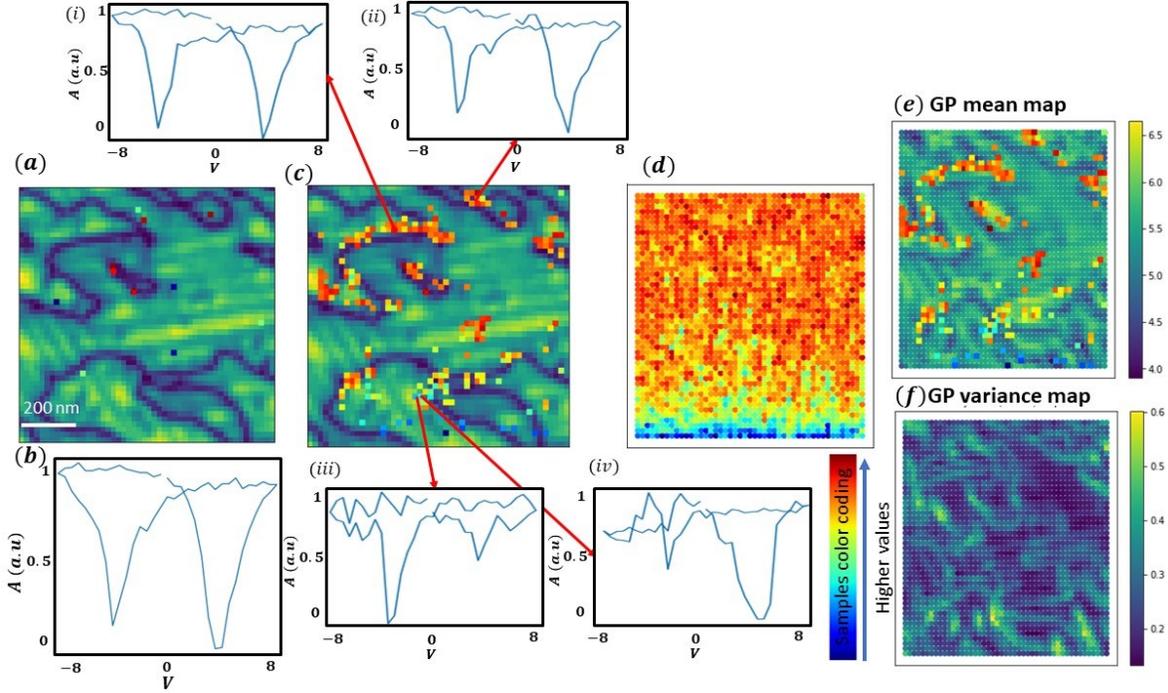

**Figure 3.** Analysis for PTO sample 2: BOARS as in Figure 2, with on a PTO sample with a different domains structure. (a) Downsampled PFM amplitude image of the PTO film, with exploration points for the spectral locations where the user voted, (b) final learned target spectral structure after voting through explored spectra in (a), (c) Plot of (a) with all the explored spectral locations overlaid (d) ground truth image, i.e., the structural similarity map as in eq. 3, given the user voted target spectral target. (e) estimated structural similarity map from the surrogate model, with all the explored spectral locations, (f) map of the model's associated uncertainty. The color of the explored locations represents the human-augmented objective function values in (a) and automated objective function values in (c), (e) as per the colormap provided. This color coding matches with the full ground truth image in (d). Within sub-figure (c), (i)-(iv) are the visualization of the spectral at some of the BO explored locations. The scale bar in (a) is 200 nm.

The above analysis shows the traditional kernel function could be unstable depending on the complexity of the parameter space and the degree of correlation between the prior knowledge (embedded in local structural image patches) and the posterior knowledge on structural similarity with the human assessed target. Our prior work[59] has shown that in such instances, it may be advantageous to utilize deep kernels in a scheme termed deep kernel learning (dKL).[48] dKL is built on the framework on fully-connected neural network (NN) where the high-dimensional input image patch is first embedded into low dimensional kernel space (in this case set as 2), and then a standard GP kernel operates, such that the parameters of GP and weights of NN are learned jointly.

This dKL technique has been implemented for better exploration through active learning in experimental environments.[4,49,60–62] Here, we utilized a DKL implementation from open-source AtomAI software package.[59]

The overall BOARS structure remains the same, but we simply replace the standard GP with a dKL-based approach. All other parameters were kept constant. The detail user voting of the spectra to set the final target for both oxide samples, similar to the analysis in figs. 2, 3, is provided in the Supplementary Material (see Figs. B2, B3). Figures 4 and 5 are the detailed analysis of the estimated spectral similarity maps, after adaptive learning with BOARS system, for first and second PTO samples respectively.

Observing both figures 4 and 5, it can be seen that the dKL method serves to better capture the correlations between the local image patches and the objective function, ultimately in adaptive learning of the estimated GP spectral similarity maps (see fig 4 (e), 5 (e)). We also observe an overall better trade-off with regards to BO exploration and exploitation, with more scattered sampling to look for potential regions of interest, particularly in fig 5 when the local structure-spectral correlation is minimal, ultimately to provide a better structural similarity map. For example, unlike in Fig. 2, the estimated uncertainty map Fig. 4 (f) within the white region has relatively lower variance, with a comparatively significant reduction of variances throughout the image space. Additionally, as in Fig. 3, BO with dKL still explores more near the phase boundary (dark channels) of the scanned image due to the input of the image patches; however, unlike the BOARS with traditional kernel, the dKL also adjusts the knowledge through posterior exploration and yields a majority of regions with high-valued targets (yellow region), as we know from the ground truth, providing a significant reduction of uncertainty as well. Thus, with the comparative analysis, we see an overall stability and enhancement of BOARS system, with efficient learning of user-desired spectral phase with incorporating local image patches of the system and rapid discovery of the changes in the structural similarity map through experimental evaluations, provided that the kernel is sufficiently expressive.

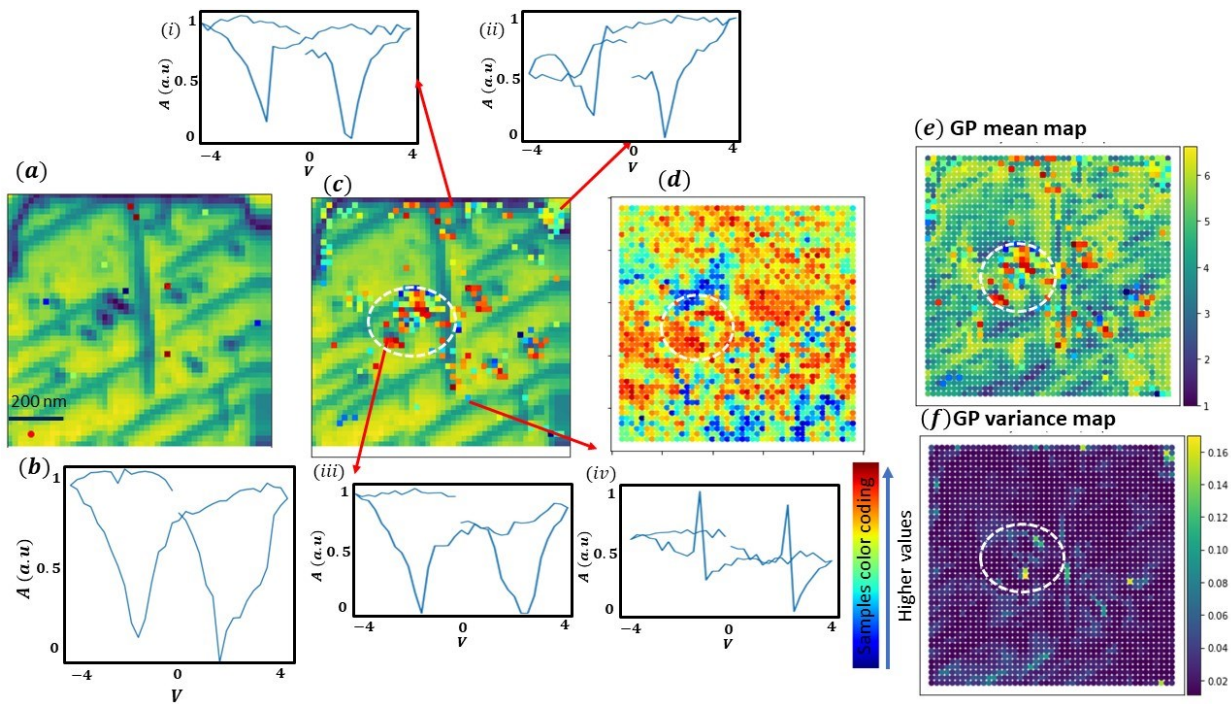

**Figure 4.** Analysis for PTO sample 1: BOARS with structural image patches and dKL kernel function. (a) Downsampled PFM amplitude image of the PTO film, with exploration points for the spectral locations where the user voted only, (b) final learned target spectral structure after voting through explored spectra in (a), (c) Plot of (a) with all the explored spectral locations overlaid (d) ground truth image, i.e., the structural similarity map as in eq. 3, given the user voted target spectral. (e) estimated structural similarity map with all the explored spectral locations, (f) uncertainty map of estimated structural similarity map. The color of the explored locations represents the human-augmented objective function values in (a) and automated objective function values in (c), (e) as per the colormap provided. This color coding matches with the full ground truth image in (d). Within sub-figure (c), (i)-(iv) are the visualization of the spectra at some of the BO explored locations.

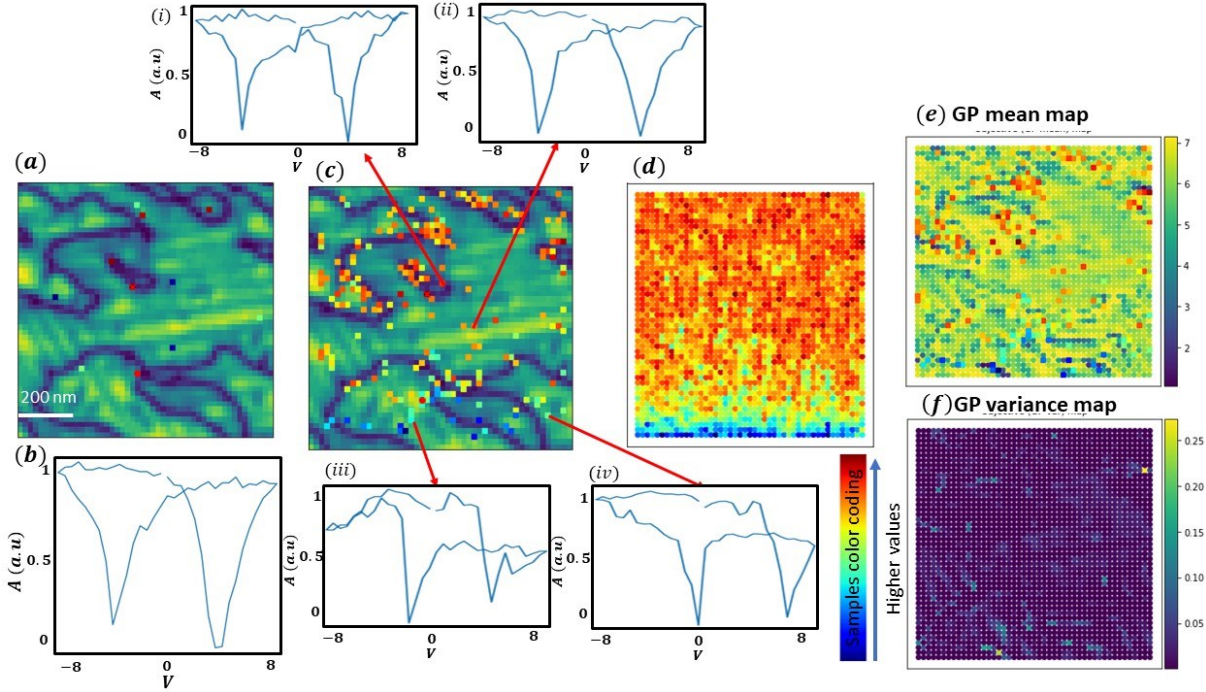

**Figure 5.** Analysis for PTO sample 2: BOARS with dKL kernel function. (a) Downsampled PFM amplitude image of the PTO film, with exploration points for the spectral locations where the user voted, (b) final learned target spectral structure after voting through explored spectra in (a), (c) Plot of (a) with all the explored spectral locations overlaid, (d) ground truth image, i.e., the structural similarity map as in eq. 3, given the user voting. (e) estimated structural similarity map from the surrogate model, with all the explored spectral locations, (f) map of the model's associated uncertainty. The color of the explored locations represents the human-augmented objective function values in (a) and automated objective function values in (c), (e) as per the colormap provided. This color coding matches with the full ground truth image in (d). Within sub-figure (c), (i)-(iv) are the visualization of the spectra at some of the BO explored locations.

To support our interpretation and validate the models, we provide the squared error map between the ground truth and the GP estimated spectral map in fig 6 for all the discussed case studies and the relative the mean squared errors (MSE) over the entire image space. For both the samples, we see an overall low MSE which shows a goodness of fit of the general BOARS architecture. For PTO sample 1, we see the MSEs are comparatively similar between the BOARS with periodic and dKL functions, with slightly better performance with dKL. However, as expected, we see significant improvement (much lower MSE) of the performance of BOARS with dKL for PTO sample 2. Furthermore, we see similar MSE values under BOARS with dKL for both the case studies which gives better stability or insensitiveness to the complexity of the problem and the efficiency of the prior knowledge (given in the form of the image patch).

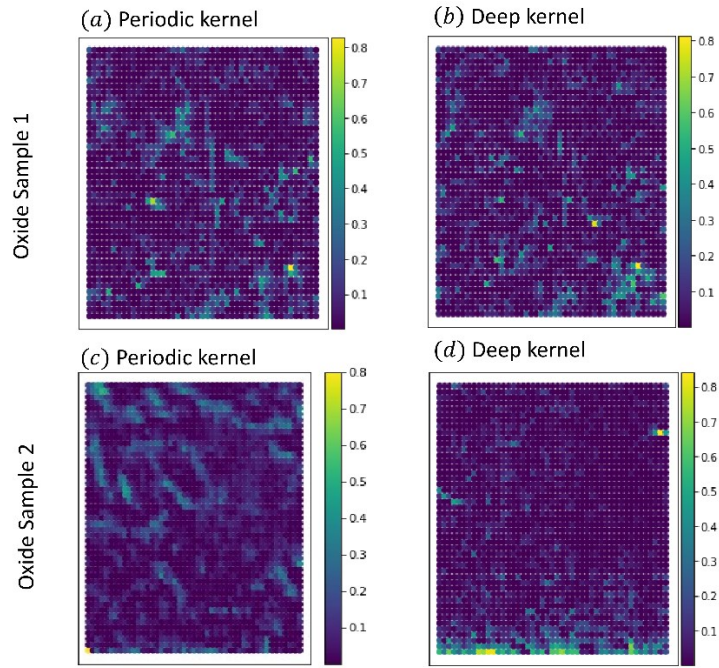

**Figure 6.** Error maps between the ground truth and the estimated spectral phase map for the following cases. (a) Oxide sample 1: BOARS with periodic kernel, (b) Oxide sample 1: BOARS with deep kernel (dKL), (c) Oxide sample 2: BOARS with periodic kernel, (d) Oxide sample 2: BOARS with deep kernel (dKL). The respective mean square errors (MSE) over the entire image space are 0.059, 0.058, 0.066 and 0.05. Note: fig (c) error map has been scaled during plotting for comparison with (d).

**Case Study: BOARS "real-time" implementation on Atomic Force Microscopy (AFM)**

After investigations on pre-acquired data, it is clear that the implementation on the real microscope will require use of deep kernel learning. As such we proceeded to apply the BOARS system with dkl kernel in real-time automated experiments on the microscope. Considering PTO sample 2, we considered the high-resolution image (128 x 128) with input image patch of window size, $w = 4\ px$. We started with 10 initial samples, $j = 10$ and 100 BO iterations.

Here also, we considered the goal to obtain a symmetrical loop, however, the voting patterns to set the target was different to our earlier analysis. This is done intentionally to understand the sensitivity of the result with different voting or targets but considering similar user-desired features. Figure 7 shows the iterative learning of the spectral structural similarity map with the BOARS system. We can see the estimated spectral similarity map (see Fig. 7(g)) show similar trends as to what we observed in Fig. 5, with a more refined map due to a higher-resolution parameter space. As in Fig. 5, we see the domain walls in the scanned image are highlighted as the potentially interesting regions of user desired spectra, and therefore the relative estimated structural similarity map has high values at the domain walls. However, as we also see from earlier analysis, the overall space is highly valued virtually throughout, and here also we see such a trend (the estimated map in Fig. 7(g) has very minimal dark regions). Regarding the computational cost, the total runtime of this AE analysis took less than 30 mins, whereas the computational cost to run experiment exhaustively for all grid points (in 128 by 128-pixel image) can take about 15-24 hours.

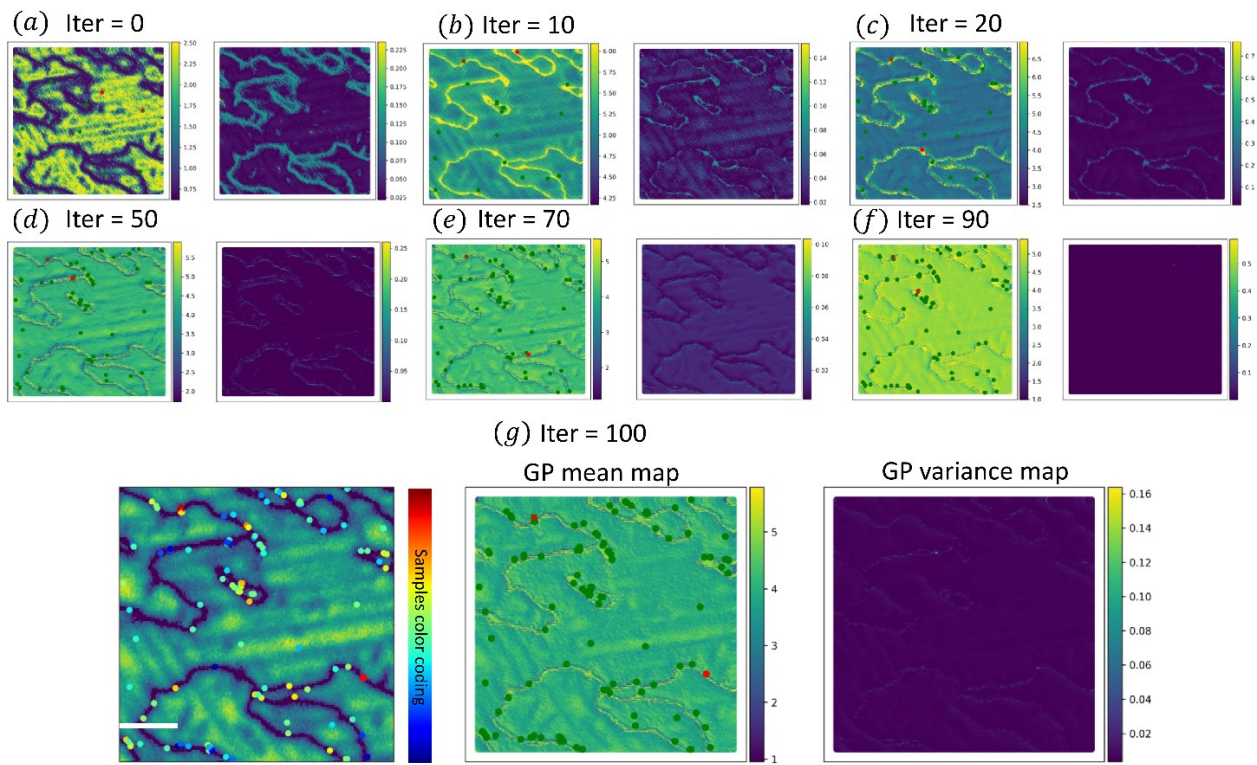

**Figure 7.** Real-time experiment and application on the microscope: Analysis for PTO sample 2: BOARS with local image patches and utilizing dKL. (a)–(f) GP estimated structural similarity map (left) and respective uncertainty map (right) for stated BO iterations. In the figures, the green dots are the explored locations while the red dots are the new locations to be explored on the next iteration. (g) analysis after BO convergence with 100 iterations. (left) high-resolution (128x128) PFM amplitude image of the PTO film, with all the explored spectral locations till BO convergence, (middle) GP prediction of structural similarity and (right) associated uncertainty map. Scale bar in (g) is 200 nm.

These results highlight two key points. One is that the degree of symmetry of the amplitude response to hysteresis loops in standard ferroelectrics like $PbTiO_3$ can be more affected by features that are not correlated with the surface domain structure, such as sub-surface defects that cannot be imaged by PFM and serve to suppress or enhance polarization. This opens the possibility to deliberately find spectral features that are not correlated with the original PFM image, and therefore, to identify notable sub-surface defect regions (for example, as in ref[63]). Secondly, the fact that the dKL method is able to learn the appropriate correlations between the local image patches and the local spectra is a key distinguishing feature. Standard kernels appear to struggle to 'ignore' the domain structure, whereas the learned kernel appears better at this task. This suggests that kernel choice is important not only for the feature learning, but for minimizing the impact of spurious correlations in active learning regimes.

**Summary**

In summary, we developed a dynamic, human-augmented Bayesian optimized active recommender system (BOARS) for curiosity-driven exploration of systems across experimental domains, where the target properties are not priorly known. The ARS system provides a framework for human in the loop automated experiments and leverages user voting as well, and a BO architecture to provide an efficient adaptive exploration towards rapid spectral learning and maximize the structural similarity of the captured spectra. We explore the effect of different kernel functions towards providing a flexible framework in a balanced learning between prior structural knowledge of local scanned image patches and the captured spectra. Future work may seek to extend the method towards multi-objective optimizations enabling users to vote based on multiple desired characteristics.

**Methods**

Table 1 provides the detailed algorithm of the BOARS system. Here we provided two objective functions formulation, based on if the user input is satisfied or not with the current target. It is to be noted the algorithm is the major contribution, specifically the human-operated process in step 2 and step 6, and therefore is the pivotal element to the paper. We described the workflow and mathematical approaches taken in Step 2 and 6 to define/update the targets and the objective functions and its connections with standard BO steps.

*Table 1: Algorithm: Bayesian optimized active recommender system (BOARS).*

1. **Segmentation of local image patches as additional channel for structure-spectra learning**:
    a. Choose a material sample. Set the control parameters of the microscope.
    b. Run microscope. Scan a high-resolution (eg. 128 by 128 grid points) image of the sample.
    c. Segment the image into several square patches with window size, $w$. The image patches are considered as input for BO, which provide the local physical information (eg. correlation) of the input location.

2. **Initialization for BO:** State maximum BO iteration, $M$. Randomly select $j$ samples (image patches), $X$. *We highlight this step as the contribution in this paper in introducing the "human operations" in the proposed AE workflow.*
    a. For sample $i$ in $j$, pass $X_i$ into microscope. Run microscope and **generate spectral data**, $S_i$.
    b. **Human-augmented process**: User votes $S_i$ with voting options, $v_i$: Bad(0), Good(1) and Very Good(2). Next follow either (c) or (d).
    c. **Generate target**: If the user voted good/very good for first time, then target, $T_i = S_i$. Normalize $T_i$.
    d. **Update target:** If $T_i \neq \emptyset$, user select preference, $p_i$ (0-1 with 1 being highest) of adding features of new spectral to the current target. Calculate $T_i$ as per eq. 1. Normalize $T_i$.

$$T_i = \left((1-p_i) * \sum_{ii=1}^{i-1} v_{ii} * T_{i-1}\right) + (p_i * v_i * S_i) \Big/ \left((1-p_i) * \sum_{ii=1}^{i-1} v_{ii}\right) + (p_i * v_i) \quad (S1)$$

   e. **Calculate human-augmented objective function**: For sample $i$ in $j$, calculate the voting augmented structural similarity index function as per eq. 2. $\psi$ is the structural similarity function; $T_j$ is the current target following step c, d, after user voted $j$ samples; $R$ is the reward parameter. $\psi$ is computed from the function "*structural_similarity*" in "*skimage.metrics*" library.

$$Y_i = \psi(T_j, S_i) + v_i * R \quad (S2)$$

   f. **Build Dataset, $D_j = \{X, Y\}$** with $X$ is a matrix with shape $(j, w*w)$ and $Y$ is an array with shape $(j)$

**Start BO.** Set $k = 1$. For $k \leq M$

3. **Surrogate Modelling**: Develop or update GPM models, given the training data, as $\Delta(D_{j+k-1})$.
    a. Optimize the hyper-parameters of kernel functions of the surrogate models.

4. **Posterior Predictions**: Given the surrogate model, compute posterior means and variances for the unexplored locations, $\overline{\overline{X_k}}$, over the parameter space as $\pi(Y(\overline{\overline{X_k}})|\Delta$ and $\sigma^2(Y(\overline{\overline{X_k}})|\Delta$ respectively.

5. **Acquisition function:** Compute and maximize acquisition function, $\max_X U(.|\Delta)$ to select next best location, $X_{j+k}$ for evaluations.

6. **Expensive Black-box evaluations:**
*We highlight this step as the contribution in this paper in introducing the "human operations" in the proposed AE workflow.*

   a. **User interaction for target update:** User gets a prompt message if the user is satisfied with the current target. User has option to choose, Yes or No. Mathematically, we can represent as $v_k = \begin{cases} 0 \ (No) \\ 1 \ (Yes) \end{cases}$
   b. **Human-augmented process:** Given $v_k = 0$, follow steps 2 (b) – 2 (e) for sample patch $X_{j+k}$. Eq. 1 and 2 can be simply modified to eq. 3 and 4 respectively.

$$T_{j+k} = \left.\left((1-p_k) * \sum_{kk=1}^{j+k-1} v_{kk} * T_{j+k-1}\right) + (p_k * v_k * S_k)\right/ \left((1-p_k) * \sum_{kk=1}^{j+k-1} v_{kk}\right) + (p_k * v_k) \quad (S3)$$

$$Y_{j+k} = \psi(T_{j+k}, S_k) + v_k * R \quad (S4)$$

   a. **Automated process:** This step is included to speed up the search process to avoid redundant user interaction in case the user is satisfied with learning of the target spectral and therefore the goal changes to learn the spectral similarity map towards achieving the converged target. Therefore, Given $v_k = 1$, $T_{j+k} = T = T_{j+k-1}$. Calculate the structural similarity index function as per eq. 5. It is to be noted that we recalculate the objective function once the user switch to human-augmented to automated process since the function changes. However, since we already have stored the previous spectral data for the explored image patches, the recalculation cost is negligible. Also, the architecture is currently set up where the switch from human-augmented to automated process is irreversible to avoid prompting user repeatedly in Step 6 (a).

$$Y_{j+k} = \psi(T, S_k) \quad (S5)$$

7. **Augmentation:** Augment data, $D_{j+k} = [D_{j+k-1}; \{X_{j+k}, Y_{j+k}\}]$.

**Supplementary Material:**

See the supplementary material for detailed descriptions and additional figures, related to the research.

**Contributions**

AB designed the algorithm, wrote the codes for implementation and analysis, analyzed data and wrote the paper. NC wrote codes for the spectral voting system. YL assisted with experimental setup and code integration. YCL and JCY grew the samples used. SJ wrote the acquisition software for python-controlled spectral acquisition. SVK assisted with analysis of results and paper writing. MZ assisted with code development in the dKL framework. RKV supervised the project, conceived of the idea, performed AFM experiments, and co-wrote the paper. All authors commented on the manuscript.

**Acknowledgements:**

The experiments, autonomous workflows and deep kernel learning was supported by the Center for Nanophase Materials Sciences (CNMS), which is a US Department of Energy, Office of Science User Facility at Oak Ridge National Laboratory. Algorithmic development was supported by the US Department of Energy, Office of Science, Office of Basic Energy Sciences, MLExchange Project, award number 107514; supported by the U.S. Department of Energy, Office of Science, Office of Basic Energy Sciences Energy Frontier Research Centers program under Award Number DE-SC0021118; and supported by University of Tennessee (Knoxville) start-up funding. J.-C.Y. and Y.-C.L. acknowledge support from National Science and Technology Council (NSTC), Taiwan, under grant no. NSTC-111-2628-M-006-005.

**Conflict of Interest:**

The authors declare no conflict of interest.

**Data Availability Statement:**

The analysis reported here is summarized in Colab Notebook for the purpose of tutorial and application to other data: https://github.com/arpanbiswas52/varTBO

Supplementary Materials of the paper titled

## "A dynamic Bayesian optimized active recommender system for curiosity-driven "Human-in-the-loop" automated experiments"

Arpan Biswas,[1,*] Yongtao Liu[1], Nicole Creange[1], Yu-Chen Liu[2], Stephen Jesse,[1] Jan-Chi Yang[2], Sergei Kalinin[3], Maxim Ziatdinov,[1,4] Rama Vasudevan[1,**]

[1] Center for Nanophase Materials Sciences, Oak Ridge National Laboratory, Oak Ridge, TN 37831

[2] Department of Physics, National Cheng Kung University, Tainan 70101, Taiwan

[3] Department of Materials Science and Engineering, University of Tennessee, Knoxville

[4] Computational Sciences and Engineering Division, Oak Ridge National Laboratory, Oak Ridge, TN 37831

**Appendix A. Gaussian Process Model (GPM)**

The general form of the GPM is as follows:

$$y(x) = x^T \beta + z(x) \quad (A.1)$$

where $x^T\beta$ is the Polynomial Regression model. The polynomial regression model captures the global trend of the data. $z(x)$ is a realization of a correlated Gaussian Process with mean $E[z(x)]$ and covariance $cov(x^i, x^j)$ functions defined as follows:

$$z(x) \sim GP\left(E[z(x)], cov(x^i, x^j)\right); \quad (A.2)$$

$$E[z(x)] = 0, cov(x^i, x^j) = \sigma^2 R(x^i, x^j) \quad (A.3)$$

$$R(x^i, x^j) = \exp\left(-0.5 * \sum_{m=1}^{d} \frac{(x_m^i - x_m^j)^2}{\theta_m^2}\right); \quad (A.4)$$

$$\theta_m = (\theta_1, \theta_2, \ldots, \theta_d)$$

where $\sigma^2$ is the overall variance parameter and $\theta_m$ is the correlation length scale parameter in dimension $m$ of $d$ dimension of $x$. These are termed as the hyper-parameters of GP model. $R(x^i, x^j)$ is the spatial correlation function. In this paper, we have considered a Radial Basis function which is given by eqn. A.4. The objective is to estimate (by MLE) the hyper-parameters $\sigma$, $\theta_m$ which creates the surrogate model that best explains the training data $D_k$ at iteration $k$.

After the GP model is fitted, the next task of the GP model is to predict at an arbitrary (unexplored) location drawn from the parameter space. Assume $\boldsymbol{D_k} = \{\boldsymbol{X_k}, Y(\boldsymbol{X_k})\}$ is the prior information from previous evaluations or experiments from high fidelity models, and $\bar{\bar{x}}_{k+1} \in \bar{\bar{X}}$ is a new design within the unexplored locations in the parameter space, $\bar{\bar{X}}$. The predictive output distribution of $x_{k+1}$, given the posterior GP model, is given by eqn A.5.

$$P(\bar{y}_{k+1}|\boldsymbol{D_k}, \bar{\bar{x}}_{k+1}, \sigma_k^2, \boldsymbol{\theta_k}) = N(\mu(\bar{y}_{k+1}(\bar{\bar{x}}_{k+1})), \sigma^2(\bar{y}_{k+1}(\bar{\bar{x}}_{k+1}))) \tag{A.5}$$

where:

$$\mu(\bar{y}_{k+1}(\bar{\bar{x}}_{k+1})) = \boldsymbol{cov}_{k+1}^T \boldsymbol{COV}_k^{-1} Y_k; \tag{A.6}$$

$$\sigma^2(\bar{y}_{k+1}(\bar{\bar{x}}_{k+1})) = cov(\bar{\bar{x}}_{k+1}, \bar{\bar{x}}_{k+1}) - \boldsymbol{cov}_{k+1}^T \boldsymbol{COV}_k^{-1} \boldsymbol{cov}_{k+1} \tag{A.7}$$

$\boldsymbol{COV_k}$ is the kernel matrix of already sampled designs $\boldsymbol{X_k}$ and $\boldsymbol{cov_{k+1}}$ is the covariance function of new design $\bar{\bar{x}}_{k+1}$ which is defined as follows:

$$\boldsymbol{COV_k} = \begin{bmatrix} cov(x_1, x_1) & \cdots & cov(x_1, x_k) \\ \vdots & \ddots & \vdots \\ cov(x_k, x_1) & \cdots & cov(x_k, x_k) \end{bmatrix}$$

$$\boldsymbol{cov_{k+1}} = [cov(\bar{\bar{x}}_{k+1}, x_1), cov(\bar{\bar{x}}_{k+1}, x_2), \ldots, cov(\bar{\bar{x}}_{k+1}, x_k)]$$

## Appendix B. Additional figures

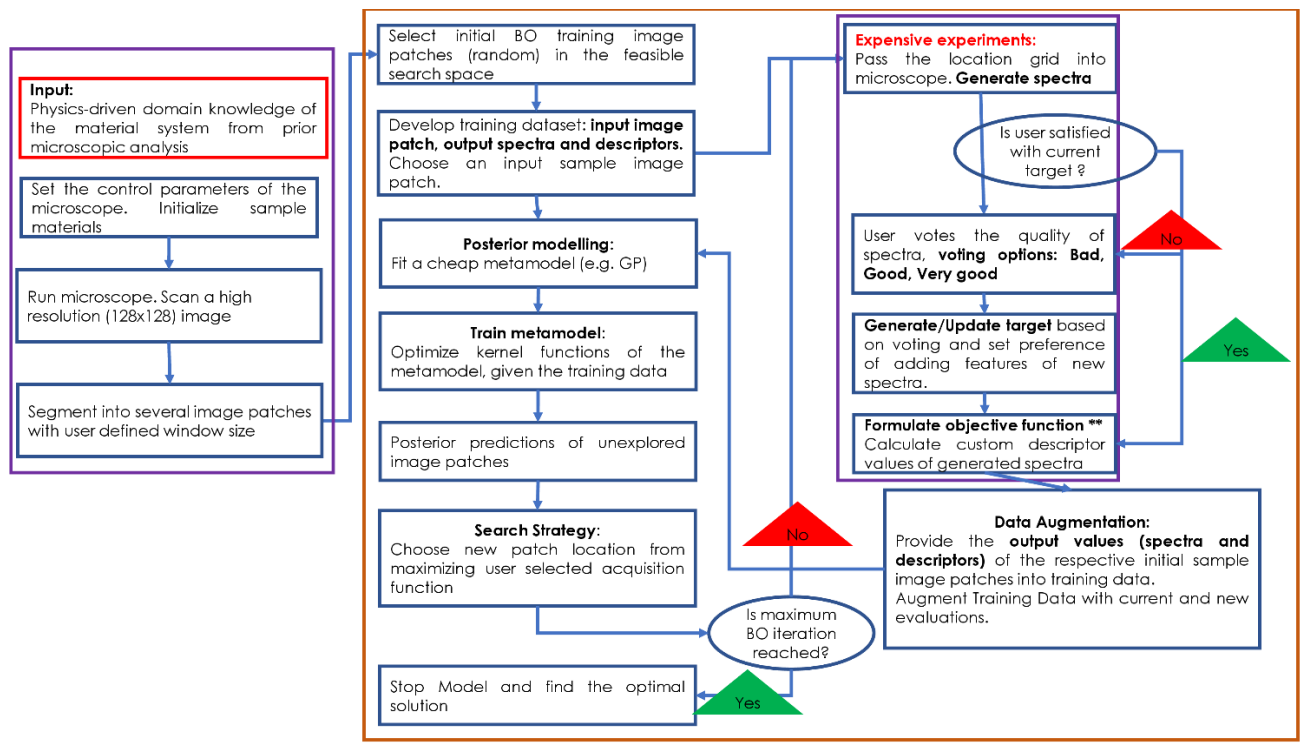

**Figure B1.** Flowchart of human augmented BOARS architecture

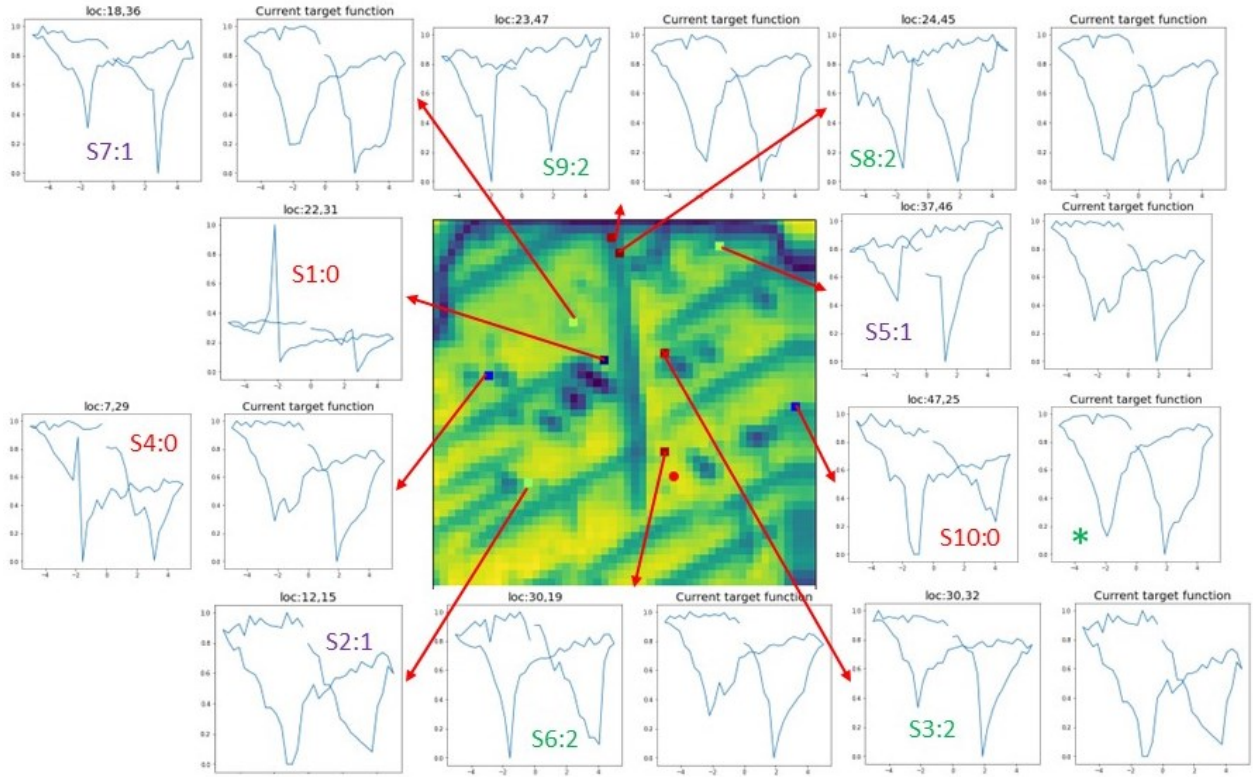

**Figure B2.** Analysis of Oxide sample 1. User-voting of the spectral at first 10 iterations. Rating for the votes is defined as: bad (0), good (1) and very good (2). We can see the dynamic changes of the target spectral in the current iteration, depending on the voting given in all the previous iterations. Finally, user satisfies with the target (represented with *) at 10th iteration. For the comparative study in this paper, this target is set for both BOARS system, with traditional periodic kernel and deep learning kernel. For all the spectral figures, the X axis ranges from -4V to 4V and the Y axis ranges from 0 A(a.u.) to 1 A(a.u).

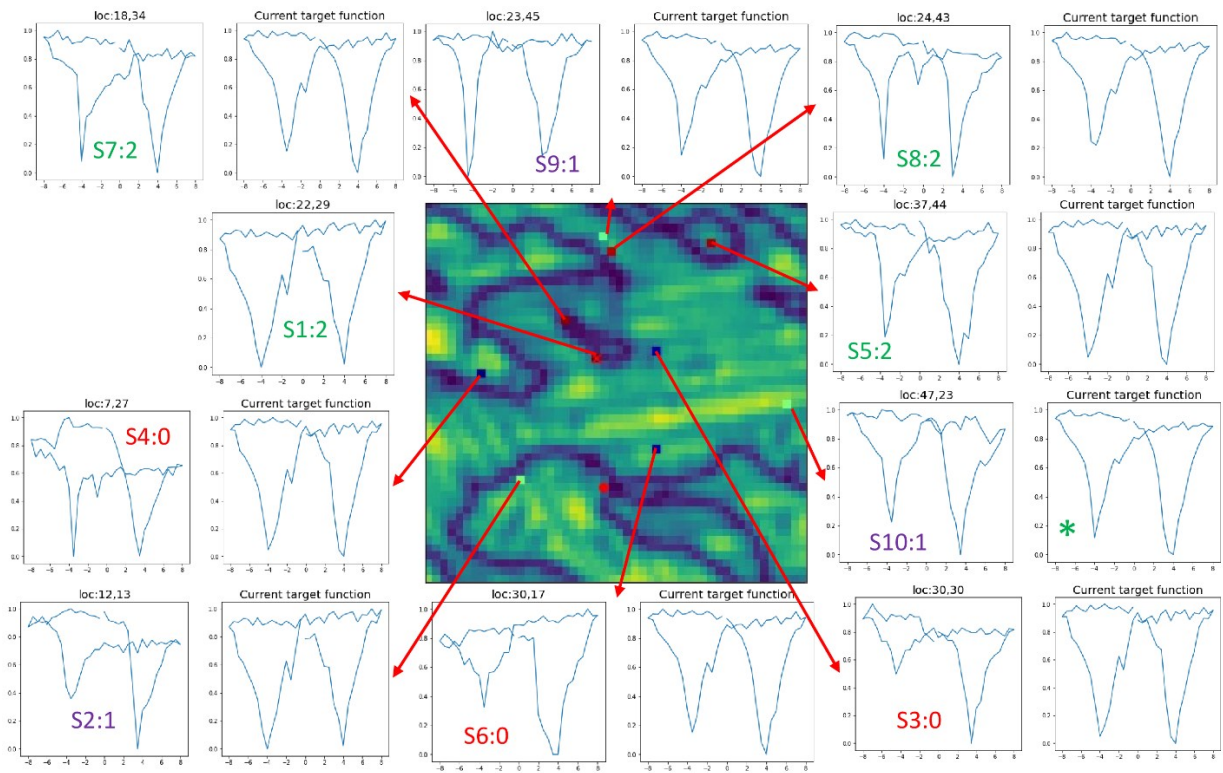

**Figure B3.** Analysis of Oxide sample 2. User-voting of the spectral at first 10 iterations. Rating for the votes is defined as: bad (0), good (1) and very good (2). We can see the dynamic changes of the target spectral in the current iteration, depending on the voting given in all the previous iterations. Finally, user satisfies with the target (represented with *) at $10^{th}$ iteration. For the comparative study in this paper, this target is set for both BOARS system, with traditional periodic kernel and deep learning kernel. For all the spectral figures, the X axis ranges from -8V to 8V and the Y axis ranges from 0 A(a.u.) to 1 A(a.u).

We can see the voting are also given based on the current target spectral structure, in other words, the votes are correlated. For example, in fig. B3, the user votes spectral 2 (S2) as good (1) whereas votes spectral 6 (S6) as bad (0), though both are spectral does not have symmetric features. This is because, with more iterations, the user is better informed about the general spectral features over the material sample and the target that can be achieved. Thus, with updated knowledge the user choice can be more specific in updating to a desired target, which is an expected human-thinking approach. In this research, we extend the application of BO to tackle this dynamic human-augmented voting functionalities in ARS system.